\documentclass{article}

\PassOptionsToPackage{numbers}{natbib}
\usepackage[final]{neurips_data_2022}
\usepackage[utf8]{inputenc} % allow utf-8 input
\usepackage[T1]{fontenc}    % use 8-bit T1 fonts
\usepackage{hyperref}       % hyperlinks
\usepackage{url}            % simple URL typesetting
\usepackage{booktabs}       % professional-quality tables
\usepackage{amsfonts}       % blackboard math symbols
\usepackage{nicefrac}       % compact symbols for 1/2, etc.
\usepackage{microtype}      % microtypography
\usepackage[dvipsnames]{xcolor}
\usepackage{soul}
\usepackage{graphicx}
\usepackage{caption}
\usepackage{subcaption}
\usepackage{multirow}
\usepackage{listings}
\usepackage[frozencache,cachedir=.]{minted}
\usepackage{tablefootnote}
\usepackage{svg}
\usepackage{amsmath}
\newminted{python}{frame=lines,framerule=2pt}

\usepackage[most]{tcolorbox}
\tcbset{
    frame code={}
    center title,
    left skip=0pt,
    left=0pt,
    leftrule=-2pt,
    rightrule=-2pt,
    right=0pt,
    top=0pt,
    bottom=0pt,
    colback=black!4,
    width=1\textwidth,
    boxsep=5pt,
    arc=0pt,outer arc=0pt,
    }
\newcommand{\thmbox}[2]{\begin{tcolorbox}#2 \end{tcolorbox}}

\newcommand{\envpool}{EnvPool}
\newcommand{\ABQ}{\texttt{ActionBufferQueue}}
\newcommand{\SBQ}{\texttt{StateBufferQueue}}
\newcommand{\TP}{\texttt{ThreadPool}}

\hypersetup{
    colorlinks=true,
    linkcolor=red,
    citecolor=blue,
    filecolor=magenta,      
    urlcolor=blue,
linktocpage}
\newcommand{\revision}[1]{#1}

\title{EnvPool: A Highly Parallel Reinforcement Learning Environment Execution Engine}

\author{
  Jiayi Weng\textsuperscript{\dag}\thanks{Currently at OpenAI. Detailed author contributions can be found in Appendix~\ref{appendix:author}.} \quad Min Lin\textsuperscript{\dag} \quad  Shengyi Huang\textsuperscript{\ddag} \quad Bo Liu\textsuperscript{$\mathsection$} \\
   \textbf{Denys Makoviichuk\textsuperscript{$\sharp$} \quad  Viktor Makoviychuk\textsuperscript{$\triangle$}\quad Zichen Liu\textsuperscript{\dag}\textsuperscript{$\square$}}\\ 
   \textbf{Yufan Song\textsuperscript{$\diamondsuit$} \quad Ting Luo\textsuperscript{$\diamondsuit$} \quad Yukun Jiang}\textsuperscript{$\diamondsuit$} \\
   \textbf{Zhongwen Xu\textsuperscript{\dag} \quad Shuicheng Yan\textsuperscript{\dag}} \\
   \\
   \textsuperscript{\dag}Sea AI Lab \\
   \textsuperscript{\ddag}Drexel University \quad \textsuperscript{\textsection}Peking University  \quad \textsuperscript{$\sharp$}Snap  \quad   \textsuperscript{$\triangle$}NVIDIA \\
   \textsuperscript{$\square$}National University of Singapore  \quad \textsuperscript{$\diamondsuit$}Carnegie Mellon University \\
  \texttt{trinkle23897@gmail.com, \{linmin,xuzw,yansc\}@sea.com }
}

\begin{document}
\maketitle
\begin{abstract}
There has been significant progress in developing reinforcement learning (RL) training systems. Past works such as IMPALA, Apex, Seed RL, Sample Factory, and others, aim to improve the system's overall throughput.
In this paper, we aim to address a common bottleneck in the RL training system, i.e., parallel environment execution, which is often the slowest part of the whole system but receives little attention. With a curated design for paralleling RL environments, we have improved the RL environment simulation speed across different hardware setups, ranging from a laptop and a modest workstation, to a high-end machine such as NVIDIA DGX-A100. On a high-end machine, EnvPool achieves one million frames per second for the environment execution on Atari environments and three million frames per second on MuJoCo environments. When running EnvPool on a laptop, the speed is 2.8$\times$ that of the Python subprocess. Moreover, great compatibility with existing RL training libraries has been demonstrated in the open-sourced community, including CleanRL, rl\_games, DeepMind Acme, etc. Finally, EnvPool allows researchers to iterate their ideas at a much faster pace and has great potential to become the de facto RL environment execution engine. Example runs show that it only takes five minutes to train agents to play Atari Pong and MuJoCo Ant on a laptop.  EnvPool is open-sourced at \url{https://github.com/sail-sg/envpool}.
\end{abstract}

\section{Introduction}

Deep Reinforcement Learning (RL) has made remarkable progress in the past years. 
Notable achievements include Deep Q-Network (DQN)~\cite{DQN}, AlphaGo~\cite{muzero,alphago,alphazero,alphago_zero}, AlphaStar~\cite{alphastar}, OpenAI Five~\cite{openai_five}, etc. 
Apart from the algorithmic innovations, the most significant improvements aimed at enhancing the training throughput for RL agents, such as leveraging the computation power of large-scale distributed systems and advanced AI chips like TPUs~\cite{TPU}.

On the other hand, academic research has been accelerated dramatically by the shortened training time. For example, DQN takes eight days and 200 million frames to train an agent to play a single Atari game~\cite{DQN}, 
 while IMPALA~\cite{IMPALA} shortens this process to a few hours and Seed RL~\cite{SeedRL} continues to push the boundary of training throughput.
This allows the researchers to perform iterations of their ideas at a much faster pace and benefits the research progress of the whole RL community.

Since training RL agents with high throughput offers important benefits, we focus on tackling a common bottleneck in the RL training system in this paper: parallel environment execution. 
To the best of our knowledge, it is often the slowest part of the whole system but has received little attention in previous research.
The inference and learning with the agent policy network can easily leverage the experience and performance optimization techniques from other areas where deep learning has been applied, like computer vision and natural language processing, often conducted with accelerators like GPUs and TPUs.
The unique technical difficulty in RL systems is the interaction between the agents and the environments.
Unlike the typical setup in supervised learning performed on a fixed dataset, the RL systems must generate environment experiences at a very fast speed to fully leverage the highly parallel computation power of accelerators.

Our contribution is to optimize the environment execution for \emph{general} RL environments, including video games and various applications of financial trading, recommendation systems, etc.
The current method to run parallel environments is to execute the environment and pre-process the observation under Python multiprocessing.
We accelerate the environment execution by implementing a general C++ threadpool-based executor engine that can run multiple environments in parallel. The well-established Python wrappers are optimized on the C++ side as well. The interactions between the agent and the environment are exposed by straightforward Python APIs as below.

\begin{minted}[frame=single,framesep=10pt]{python}
import envpool
import numpy as np

# make gym env
env = envpool.make("Pong-v5", env_type="gym", num_envs=100)
obs = env.reset()  # with shape (100, 4, 84, 84)
act = np.zeros(100, dtype=int)
obs, rew, done, info = env.step(act, env_id=np.arange(100))
# can get the next round env_id through info["env_id"]
\end{minted}

The system is called \envpool, a highly parallel reinforcement learning environment execution engine, where we support OpenAI \texttt{gym} APIs and DeepMind \texttt{dm\_env} APIs. \envpool~has both synchronous and asynchronous execution modes, where the latter is rarely explored in the mainstream RL system implementation even though it has enormous potential. 
The currently supported environments on \envpool~include Atari~\cite{ALE}, MuJoCo~\cite{MuJoCo}, DeepMind Control Suite~\cite{dmc}, ViZDoom~\cite{VizDoom}, classic RL environments like mountain car, cartpole~\cite{RLbook}, etc. 

There are two groups of targeted users for \envpool. One is RL researchers and practitioners who do not have to modify any parts of the RL environments. For example, researchers who would like to train an agent on Atari / MuJoCo tasks. They can use \envpool~just as OpenAI Gym, but faster. \envpool~intends to cover as many standard RL environments as possible in our GitHub repository. This group of users does not need to understand any internals of \envpool, including any C++ code. They only work with the Python APIs (See Appendix~\ref{appendix:python_apis} for comprehensive user APIs). The second group of ``users'' which we would like to call developers, are familiar with RL environment implementation (in C++) and would like to integrate their loved RL environments into \envpool~to speed up the environment execution. For this developer's group, we have provided extensive documentation on integrating a C++-based RL environment into \envpool, including some straightforward examples (See Section~\ref{section:methods} for more technical details).

Performance highlights of the \envpool~system include:

\begin{itemize}
\item With 256 CPU cores on an NVIDIA DGX-A100 machine, \envpool~achieves a simulation throughput of one million frames per second on Atari and three million physics steps per second on MuJoCo environments, which is 14.9$\times$ / 19.2$\times$ improvement over the current popular Python implementation~\cite{gym} (i.e., 72K frames per second / 163K physics steps per second for the same hardware setup).

\item On a laptop with 12 CPU cores, \envpool~obtains a speed 2.8$\times$ of the Python implementation.

\item When integrated with existing RL training libraries, example runs show that we can train agents to play Atari Pong and MuJoCo Ant on a laptop in five minutes.

\item Sample efficiency is not sacrificed when replacing OpenAI gym with EnvPool and keeping the same experiment configuration. It is a pure speedup without cost.

\end{itemize}

\section{Related Works}

In this section, we review the existing RL environment execution component in the literature. 
Most implementations in RL systems use Python-level parallelization, e.g., For-loop or subprocess~\cite{gym}, in which we can easily run multiple environments and obtain the interaction experience in a batch.
While the straightforward Python approaches are plugged easily with existing Python libraries and thus widely adopted, they are computationally inefficient compared to using a C++-level thread pool to execute the environments. 
The direct outcome of using inefficient environment parallelization is that more machines have to be used just for environment execution. 
Researchers build distributed systems like Ray~\cite{Ray} which allow easy distributed remote environment execution. 
Unfortunately, multiple third parties report an inefficient scale-up experience using Ray RLlib~\cite{RLlib,Ray} (cf. Figure 3 in \cite{sample-factory}).
This issue might be because, in a distributed setup, Ray and RLlib have to trade-off the communication costs with other components and are not specifically optimized for environment execution. 

Sample Factory~\cite{sample-factory} focuses on optimizing the entire RL system for a single-machine setup instead of a distributed computing architecture. 
To achieve high throughput in the action sampling component, they introduce a sophisticated, fully asynchronous approach called Double-Buffered Sampling, which allows network forwarding and environment execution to run in parallel but on different subsets of the environments.
Though having improved the overall throughput dramatically over other systems, the implementation complexity is high, and it is not a standalone component that can be plugged into other RL systems.
Furthermore, Sample Factory sacrifices compatibility with a family of RL algorithms that can only work in synchronous mode to achieve high throughput.
In contrast, \envpool~has both properties of high throughput and great compatibility with existing APIs and RL algorithms.

A few recent works, e.g., Brax~\cite{brax}, Isaac Gym~\cite{isaac-gym}, and WarpDrive~\cite{warpdrive}, use accelerators like GPUs and TPUs for the environment engine.
Due to the highly parallel nature of the accelerators, numerous environments can be executed simultaneously. The intrinsic drawback of this approach is that the environments must be purely compute-based, i.e., matrix operations so that they can be easily accelerated on GPUs and TPUs. They cannot handle general environments like common video games, e.g., Atari~\cite{ALE}, ViZDoom~\cite{VizDoom}, StarCraft~II~\cite{alphastar}, and Dota~2~\cite{openai_five}. Moreover, in real-world applications, most scenarios cannot be converted into a pure compute-based simulation. Such a major drawback places the applications of this approach on a very limited spectrum.

The most relevant work to ours is the PodRacer architecture~\cite{podracer}, which also implements the C++ batched environment interface and can be utilized to run general environments. 
However, their implementation only supports synchronous execution mode where PodRacer is operated on the whole set of environments at each timestep. 
The stepping will wait for the results returned by all environments and thus be slowed down significantly by the slowest single environment instance.
The description of PodRacer architecture is specific to the TPU configuration. 
PodRacer is not open-sourced, and we cannot find many details on the concrete implementation.
In contrast, \envpool~uses the asynchronous execution mode as a default to avoid slowing down due to any single environment instance. 
Moreover, it is not tied to any specific computing architectures. 
We have run \envpool~on both GPU machines and Cloud TPUs.

\section{Methods}
\label{section:methods}
\thmbox{note}{This section is largely intended for developers who are interested in the technical details of EnvPool and would like to contribute to the community. See Appendix~\ref{appendix:python_apis}~for detailed usage of \envpool~for RL researchers and practioners.}

\envpool~contains three key components optimized in C++, the \texttt{ActionBufferQueue}, the \texttt{ThreadPool} and the \texttt{StateBufferQueue}. It uses pybind11~\cite{pybind11} to expose the user interface to Python. In this section, we start by providing an overview of the overall system architecture. We then illustrate the optimizations we made in the individual components. Finally, we briefly describe how to add new RL environments into EnvPool. Complete Python user APIs can be found in Appendix~\ref{appendix:python_apis}.

\subsection{Overview}

\begin{figure}[tbp]
    \centering
    \includegraphics[width=\linewidth]{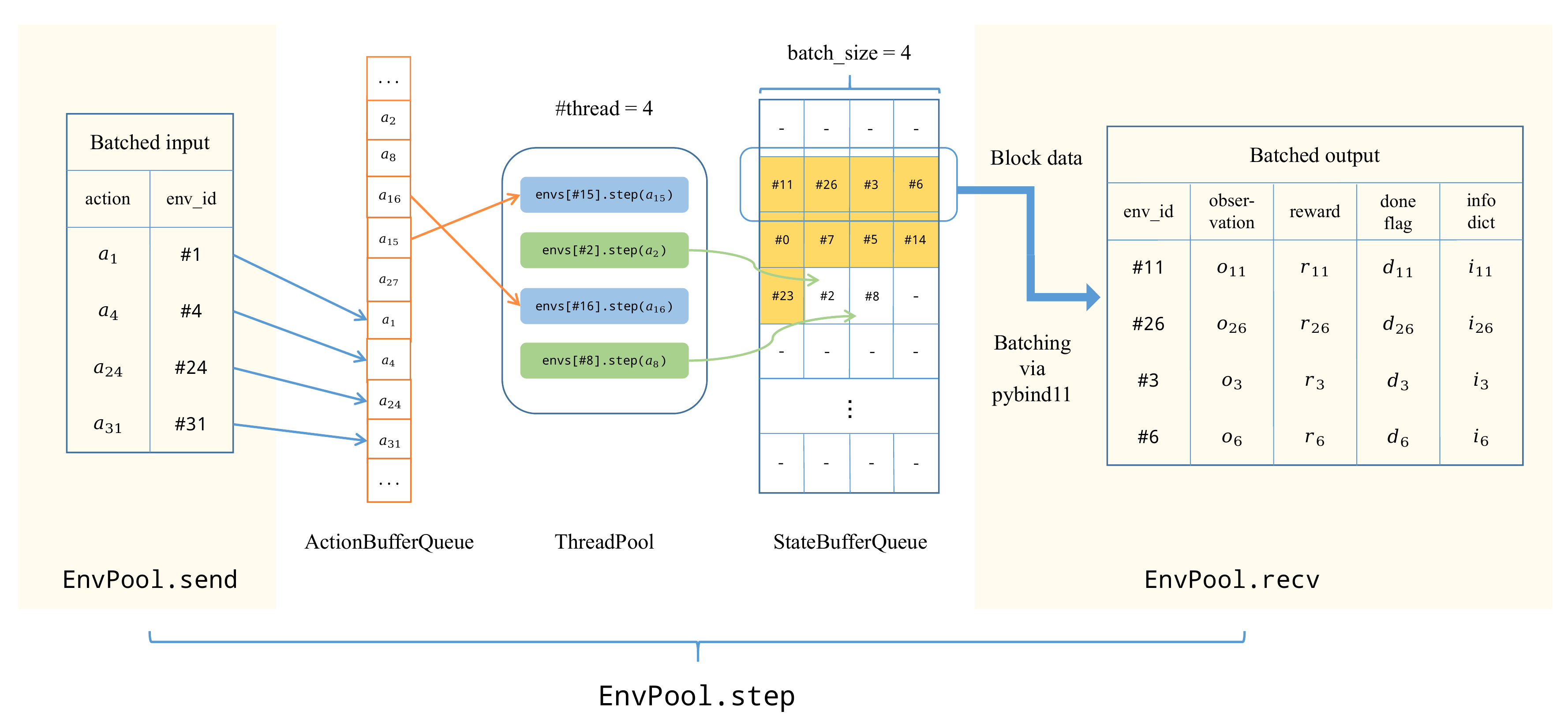}
    \caption{\envpool~System Overview}
    \label{fig:sys}
\end{figure}

In the APIs of \texttt{gym} and \texttt{dm\_env}, the central way to interact with RL environments is through the \texttt{step} function. The RL agent sends an action to the environment, which returns an observation. To increase the throughput of this interaction, the typical approach is to replicate it in multiple threads or processes. However, in systems that prioritize throughput (such as web services), the asynchronous event-driven pattern often achieves better overall throughput. This is because it avoids the context switching costs that arise in a simple multi-threaded setting.

EnvPool follows the asynchronous event-driven pattern visualized in Figure~\ref{fig:sys}. Instead of providing a synchronous \texttt{step} function, in each interaction, \envpool~receives a batched action through the \texttt{send} function. The \texttt{send} function only puts these actions in the \texttt{ActionBufferQueue}, and returns immediately without waiting for environment execution. Independently, threads in the \texttt{ThreadPool} take action from the \texttt{ActionBufferQueue} and perform the corresponding environment execution. The execution result is then added to the \texttt{StateBufferQueue}, which is pre-allocated as blocks. A block in \texttt{StateBufferQueue} contains a fixed number (\texttt{batch\_size} in the next section) of states. Once a block in the \texttt{StateBufferQueue} is filled with data, \envpool~will pack them into NumPy~\cite{numpy} arrays. The RL algorithm receives a batch of states by taking from the \texttt{StateBufferQueue} via the \texttt{recv} function. Details on the \texttt{ActionBufferQueue} and \texttt{StateBufferQueue} can be found in Appendix~\ref{appendix:buffers}.

A traditional \texttt{step} can be seen as consequent calls to \texttt{send} and \texttt{recv} with a single environment. However, separating \texttt{step} into \texttt{send/recv} provides more flexibility and opportunity for further optimization, e.g., they can be executed in different Python threads.

\subsection{Synchronous vs. Asynchronous}

\begin{figure}[htp]
    \centering
    \includegraphics[width=0.6\linewidth]{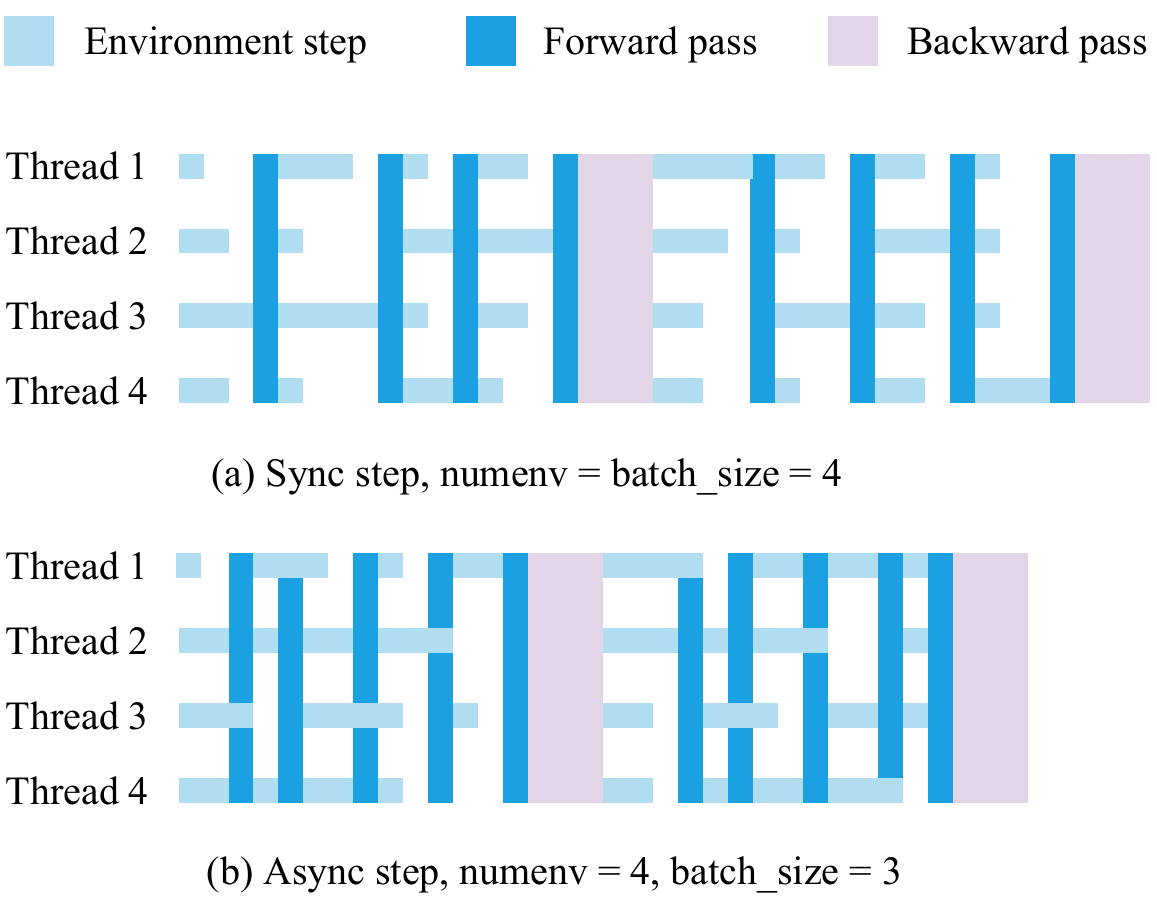}
    \caption{Synchronous step vs asynchronous step in \envpool.}
    \label{fig:async}
\end{figure}

A popular implementation of vectorized environments like \texttt{gym.vector\_env}~\cite{gym} executes all environments synchronously in the worker threads. We denote the number of environments \texttt{num\_envs} as $N$. In each iteration, the input $N$ actions are first distributed to the corresponding environments, then wait for all $N$ environments to finish their executions. The RL agent will receive $N$ observation arrays and predict $N$ actions via forward pass. As shown in Figure~\ref{fig:async}~(a), the performance of the synchronous step is determined by the slowest environment execution time, and hence not efficient for scaling out.

%\begin{figure}[tbp]

Here we introduce a new concept \texttt{batch\_size} in \envpool's asynchronous \texttt{send/recv} execution. This idea was first proposed by Tianshou~\cite{tianshou}. \texttt{batch\_size} (denoted as $M$) is the batch size of environment outputs expected to be received by \texttt{recv}. As such, \texttt{batch\_size}~$M$ cannot be greater than \texttt{num\_envs} $N$.

In each iteration, \envpool~only waits for the outputs of the first $M$ environment steps, and let other (unfinished) thread executions continue at the backend. Figure~\ref{fig:async}~(b) demonstrates this process with $N=4$ and $M=3$ in 4 threads. Compared with a synchronous step, asynchronous \texttt{send/recv} has a considerable advantage when the environment execution time has a large variance, which is a common when $N$ is large.

\envpool~can switch between synchronous mode and asynchronous mode by only specifying different \texttt{num\_envs} and \texttt{batch\_size}. In the asynchronous mode, \texttt{batch\_size < num\_envs}, the throughput is maximized. To switch to synchronous mode, we only need to set \texttt{num\_envs = batch\_size}, then consecutive calling \texttt{send/recv} is equivalent to synchronously stepping all the environments.

\subsection{ThreadPool}

ThreadPool~\cite{threadpool} is a multi-thread executor implemented with \texttt{std::thread}. It maintains a fixed number of threads waiting for task execution without creating or destroying threads for short-term tasks. ThreadPool is designed with the following considerations:

\begin{itemize}
    \item To minimize context switch overhead, the number of threads in ThreadPool is usually limited by the number of CPU cores. 
    \item To further speed up ThreadPool execution, we can pin each thread to a pre-determined CPU core. This further reduces context switching and improves cache efficiency.
    \item We recommend setting \texttt{num\_env} $N$ to be 2$-$3$\times$ greater than the number of threads to keep the threads fully loaded when part of the envs are waiting to be consumed by the RL algorithm. On one hand, if we treat the environment execution time as a distribution, taking the $M$ environments with the shortest execution times can effectively avoid the long-tail problem; on the other hand, adding too many environments but keeping the \texttt{batch\_size} unchanged may cause sample inefficiency or low-utilization of computational resources.
\end{itemize}

\subsection{Adding New RL Environments}

EnvPool is a highly extensible and developer-friendly platform for adding new reinforcement learning environments. The process is well-documented and straightforward for C++ developers\footnote{\url{https://envpool.readthedocs.io/en/latest/content/new_env.html}}.

First, developers need to implement the RL environment in a C++ header file. This involves defining the \texttt{EnvSpec} and the environment interface, which includes methods like \texttt{Reset}, \texttt{Step}, and \texttt{IsDone}. Next, they need to write a Bazel BUILD file to manage dependencies. They can then use these C++ source files to generate a dynamically linked binary, which can be instantiated in Python using pybind11. Finally, they need to register the environment in Python side and write rigorous unit tests for debugging.

One of the advantages of EnvPool is that adding new RL environments does not require a deep understanding of the core infrastructure. This makes it easy for developers to experiment with different environments and push the boundaries of RL research.

\section{Experiments}
Our experiments are divided into two parts. In the first part, we evaluate the simulation performance of the reinforcement learning environment execution engines, using randomly sampled actions as inputs. This isolated benchmark allows us to measure the performance of the engine component without the added complexity of agent policy network inference and learning.

In the second part of our experiments, we assess the impact of using \envpool~with existing RL training frameworks. We test \envpool~with CleanRL~\cite{huang2021cleanrl}, rl\_games~\cite{rl-games2022}, and DeepMind's Acme framework~\cite{acme} to see how it can improve overall training performance.

Overall, our experiments demonstrate the value of \envpool~as a tool for improving the efficiency and scalability of RL research. By optimizing the simulation performance of RL environments, \envpool~allows researchers to train agents more quickly and effectively.

\subsection{Pure Environment Simulation}

\begin{figure}[t]
    \centering
    \includegraphics[width=1\linewidth]{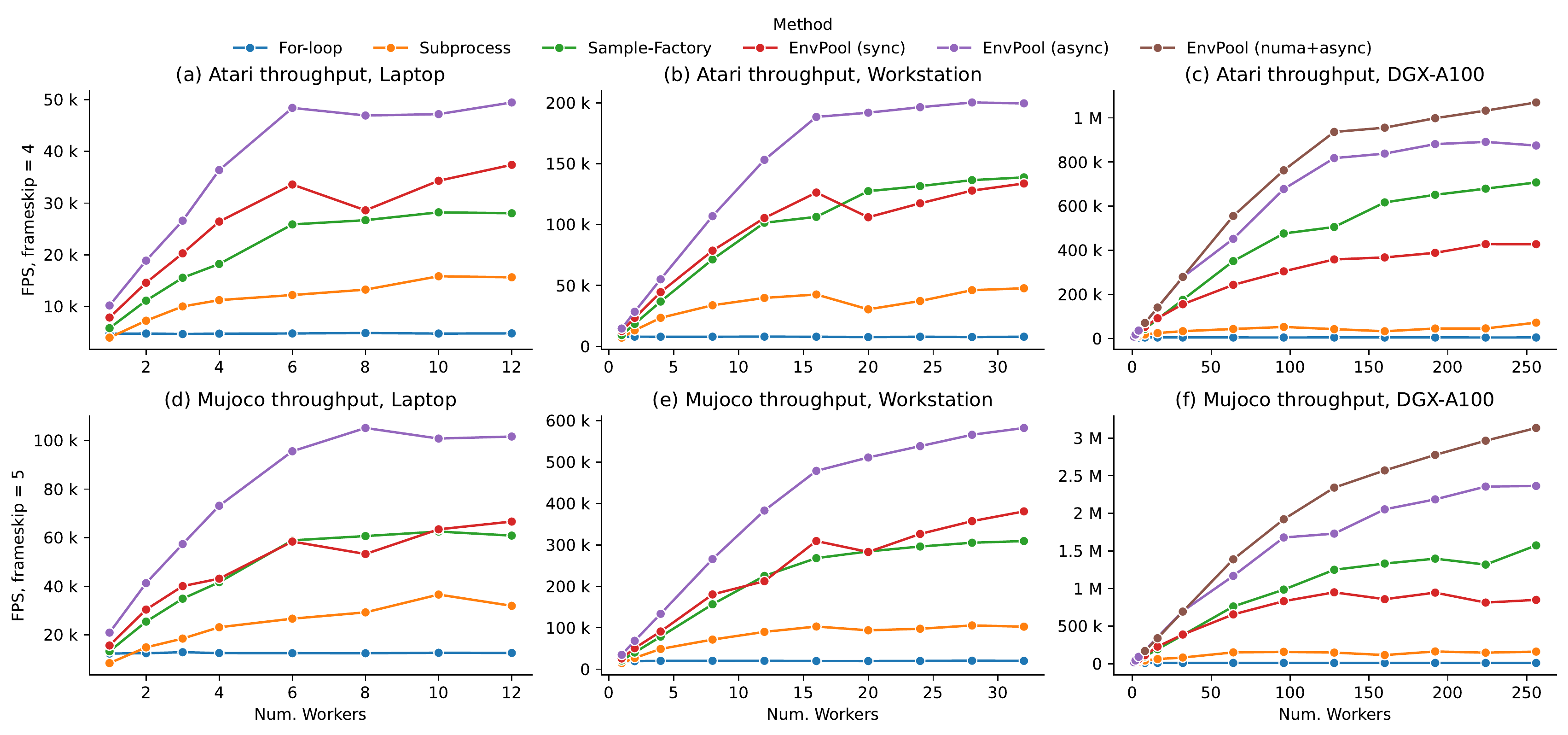}
    \caption{Simulation throughput in three machines with Atari and MuJoCo tasks.}
    \label{fig:benchmark}
\end{figure}

We first evaluate the performance of \envpool~against a set of established baselines on the RL environment execution component. 
Three hardware setups are used for the benchmark: the Laptop setting, the Workstation setting, and the NVIDIA DGX-A100 setting. Detailed CPU types and specifications can be found in Appendix~\ref{appendix:cpu_type}.

The laptop has 12 CPU cores, and the workstation has 32 CPU cores. Evaluating \envpool~on these two configurations can demonstrate its effectiveness with small-scale experiments. An NVIDIA DGX-A100 has 256 CPU cores with 8 NUMA nodes. Note that running multi-processing on each NUMA node not only makes the memory closer to the processor but also reduces the thread contention on the \ABQ.

As for the RL environments, we experiment on two of the most used RL benchmarks, namely, Atari~\cite{ALE} with Pong and MuJoCo~\cite{MuJoCo} with Ant.
In the experiments of pure environment simulation, we obtain a randomly sampled action based on the action space definition of the games and send the actions to the environment executors. The number of frames per second is measured with a mean of 50K iterations, where the Atari frame numbers follow the practice of IMPALA~\cite{IMPALA} and Seed RL~\cite{SeedRL} with frameskip set to 4, and MuJoCo sub-step numbers set to 5.

We compare several concrete implementations extensively, which are described below. Among them, Subprocess is the most popular implementation currently and, to the best of our knowledge, Sample Factory is the best performing general RL environment execution engine at the time of publication. 
\begin{itemize}
    \item For-loop: execute all environment steps synchronously within only one thread;
    \item Subprocess~\cite{gym}: execute all environment steps synchronously with shared memory and multiple processes.
    \item Sample Factory~\cite{sample-factory}: pure asynchronous step with a given number of worker threads; we pick the best performance over various  \texttt{num\_envs} per worker.
    \item \envpool~(sync): synchronous step execution in \envpool.
    \item \envpool~(async): asynchronous step execution in \envpool; given several worker threads for \texttt{batch\_size}, pick the best performance over various \texttt{num\_envs}.
    \item \envpool~(numa+async): use all NUMA nodes, each launches \envpool~individually with asynchronous execution to see the best performance of \envpool.
\end{itemize}

To demonstrate the scalability of the above methods, we conduct experiments using various numbers of workers for the RL environment execution. The experiment setup ranges from a couple of workers (e.g., 4 cores) to using all the CPU cores in the machine (e.g., 256 cores).

\begin{table}[t]
\centering
\small
\caption{Numeric results for benchmarking.}
\label{table:benchmark}
\begin{tabular}{l|cc|cc|cc}
\toprule
System   Configuration                       & \multicolumn{2}{c}{Laptop} & \multicolumn{2}{|c}{Workstation} &   \multicolumn{2}{|c}{DGX-A100} \\\midrule
Method \textbackslash~Env (FPS) & Atari        & MuJoCo     & Atari         & MuJoCo         & Atari & MuJoCo \\
\midrule
For-loop & 4,893 & 12,861 & 7,914 & 20,298 &  4,640 & 11,569 \\
Subprocess & 15,863 & 36,586 & 47,699 & 105,432 &  71,943 & 163,656 \\
Sample-Factory & 28,216 & 62,510 & 138,847 & 309,264 &  707,494 & 1,573,262 \\\midrule
\envpool~(sync) & 37,396 & 66,622 & 133,824 & 380,950 & 427,851 & 949,787 \\
\envpool~(async) & \textbf{49,439} & \textbf{105,126} & \textbf{200,428} & \textbf{582,446} & 891,286 & 2,363,864 \\
\envpool~(numa+async) & / & / & / & / & \textbf{1,069,922} & \textbf{3,134,287} \\\bottomrule
\end{tabular}
\end{table}

Our \envpool~system outperforms all of the strong baselines with significant margins on all hardware setups of the Laptop, Workstation, and DGX-A100 (Figure~\ref{fig:benchmark} and Table~\ref{table:benchmark}). 
The most popular Subprocess implementation has extremely poor scalability with an almost flat curve. 
This indicates a small improvement in throughput with the increased number of workers and CPUs. The poor scaling performance of Python-based parallel execution confirms the motivation of our proposed solution.

The second important conclusion is that, even if we use a single environment in \envpool, we can get a free $\sim$2$\times$ speedup. Complete benchmarks on Atari, MuJoCo, and DeepMind Control can be found in Appendix~\ref{appendix:single_env}.

The third observation is that synchronous modes have significant performance disadvantages against asynchronous systems. 
This is because the throughput of the synchronous mode execution is determined by the slowest single environment instance, where the stepping time for each environment instance may vary considerably, especially when there is a large number of environments.

\subsection{End-to-end Agent Training}
\label{sec:end-to-end-training}

In this work, we demonstrate successful integration of EnvPool into three different open-sourced RL libraries. 
EnvPool can serve as a drop-in replacement of the vectorized environments in many deep RL libraries and reduce the training wall time without sacrificing sample efficiency. The integration with training libraries has been straightforward due to compatibility with existing environment APIs. These example runs were performed by practitioners and researchers themselves, reflecting realistic use cases (e.g., using their machines and their preferred training libraries) in the community. 

The full results cover a wide range of combinations to demonstrate the general improvement on different setups, including different training libraries (e.g., PyTorch-based, JAX-based), RL environments (e.g., Atari, MuJoCo), machines (e.g., laptops, TPU VMs). We present the main findings in the following paragraphs, \revision{where results are aggregated over five random seeds, the learning curves are smoothed by a moving average of window size 10, and the shaded region of the learning curves represents one standard deviation of episodic returns.} The complete configurations and results can be found in Appendix~\ref{appendix:end_to_end}. Note that the hardware specifications of these experiments are different thus readers should \emph{not} compare training speeds across different training libraries.

\begin{figure}[t]
    \centering
    \includegraphics[width=1\linewidth]{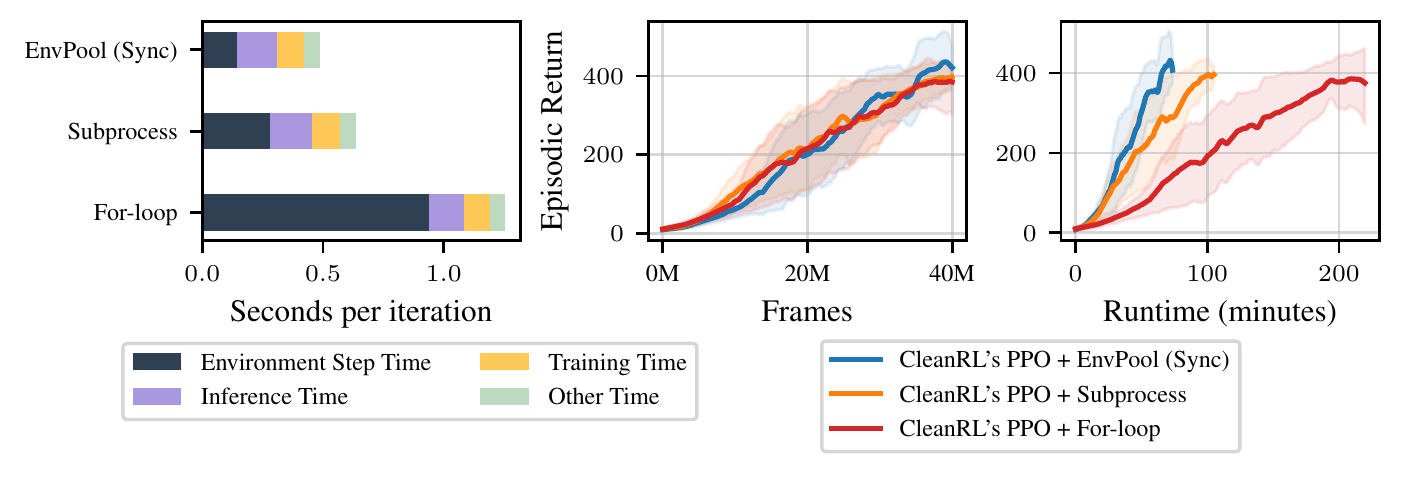}
        \caption{A profile of CleanRL's PPO in the Atari game Breakout using $N=8$.}
                \label{fig:case-study}
\end{figure}

\revision{\textbf{How much time does parallel environment execution take?} As a case study, we profile CleanRL's PPO in Atari games with three parallelization paradigms — For-loop, Subprocess, and \envpool~(Sync). CleanRL's PPO is empirically shown to be equivalent to \texttt{openai/baselines}’ PPO \citep{shengyi2022the37implementation}, and we use the same PPO hyperparameters used in the original PPO paper~\citep{PPO}, which uses $N=8$.  Specifically, we measure the following times per iteration over 9,765 iterations of rollout and training:
\begin{enumerate}
    \item Environment Step Time: the time spent on \texttt{env.step(act)} (i.e., stepping 8 actions in 8 environments and returning a batch of 8 observations, rewards, dones, and infos).
    \item Inference Time: the time spent on computing actions, log probabilities, values, and the entropy.
    \item Training Time: the time spent doing forward and backward passes.
    \item Other Time: the time spent on other procedures (e.g., storage, moving data between GPU and CPU, writing metrics, etc).
\end{enumerate}
The results are presented in Figure~\ref{fig:case-study}. We clearly see that Environment Step Time is a significant bottleneck under Python-level parallelization, and EnvPool (Sync) ameliorates this bottleneck. As a result, the end-to-end training time decreased from ~200 minutes (CleanRL's PPO + For-loop) to approximately 73 minutes (CleanRL's PPO + EnvPool (Sync)) while maintaining sample efficiency. Furthermore, we should expect a further speed up with a larger number of $N$, such as 32 or 64 when using EnvPool compared to other parallelization paradigms.}

\revision{Note that the above case study could look drastically different based on 1) the number of environments (e.g., $N=8$, $32$, or $64$); 2) the type of environments (e.g., MuJoCo or Atari); 3) the learning parameters (e.g., the number of mini batches used); and 4) the used deep RL library. Even so, what is important is that \envpool~ can speed up the Environment Step Time in the overall training system with $N>1$.}

\begin{figure}[t]
    \centering
    \includegraphics[width=1\linewidth]{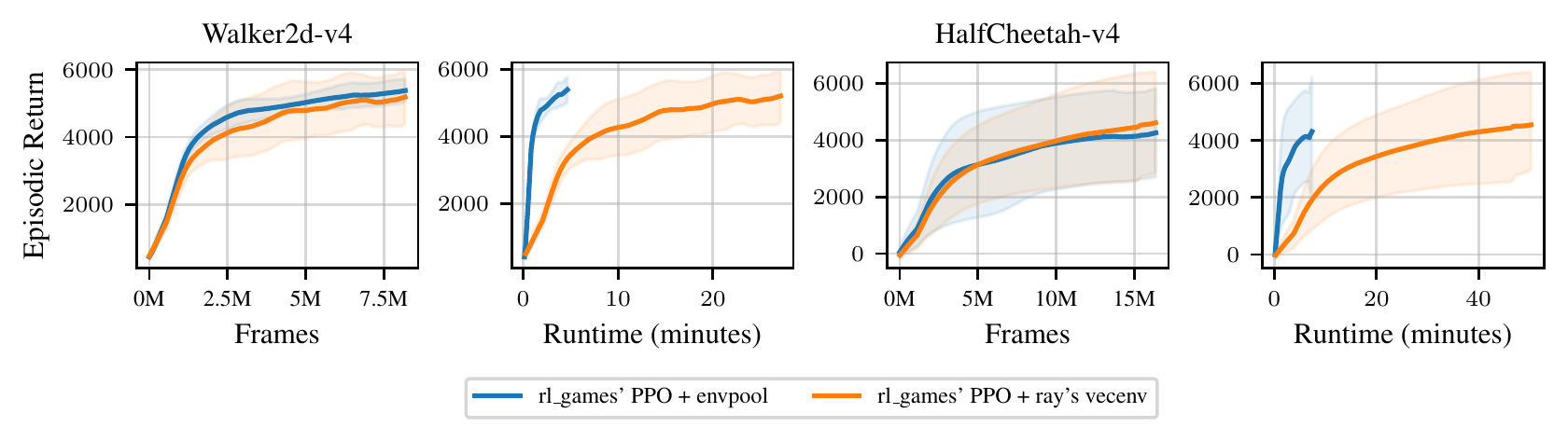}    
        \caption{rl\_games example runs with Ray and with EnvPool, using the same number of parallel environments $N=64$}
                \label{fig:walker2d_same_N}
\end{figure}

\revision{
\textbf{Easy integration with popular deep RL libraries.} Since many deep RL libraries utilize vectorized environments with some form of parallel environment executors, integrating EnvPool to them is straightfoward. In this work, we additionally present successful integration with rl\_games~\cite{rl-games2022} and DeepMind Acme~\citep{acme}. For example, Figure~\ref{fig:walker2d_same_N} shows multiple folds of wall-time training speed improvement in rl\_games when using EnvPool versus its default parallel environment executor built on top of Ray~\cite{Ray}. Further results on Acme can be found in Appendix~\ref{appendix:end_to_end}.}

\textbf{High throughput training.} Additionally, we can search for an alternative set of hyperparameters that better leverage EnvPool's throughput. \revision{For example, in MuJoCo, Schulman et al.~\cite{PPO} use a single simulation environment and let PPO use 32 mini-batches and 10 update epochs, which results in 320 gradient updates per batch of rollout data. This results in stale data after the first gradient update (i.e., the optimized policy is newer than the behavior policy that was used to collect the rollout data; see \citep{shengyi2022the37implementation} for more details). To reduce the stale data, we could use a higher number of simulation environments, such as $N=64$ and fewer mini-batches and update epochs.}

\begin{figure}[t]
    \centering
    \includegraphics[width=1\linewidth]{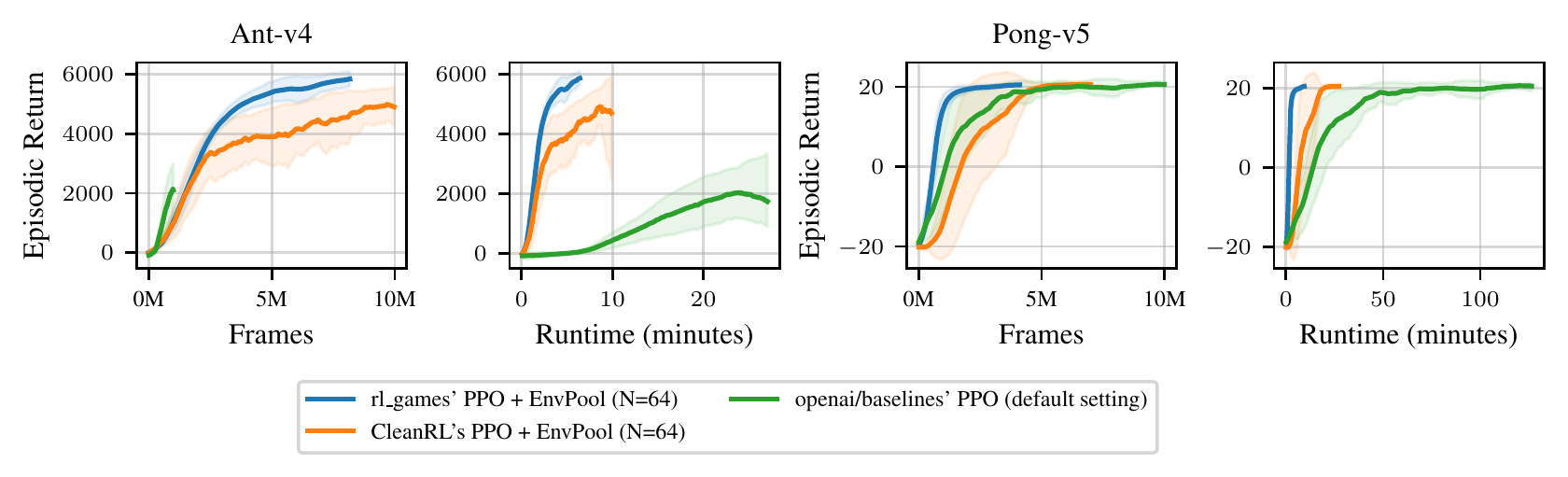}    
        \caption{rl\_games and CleanRL example runs with $N=64$ and tuned parameters compared to \texttt{openai/baselines}' PPO which by default uses $N=1$ for MuJoCo and $N=8$ for Atari experiments~\cite{PPO}.}
                \label{fig:baselines-comparison}
\end{figure}

For example, in Figure~\ref{fig:baselines-comparison} example runs, rl\_games PPO can solve Ant in five minutes of training time, while OpenAI baselines' PPO can only get to score 2,000 in 20 minutes. Such a significant speed up on a laptop-level machine benefits researchers in terms of a rapid turnaround time of their experiments. We note that a drop in sample efficiency is observed in these runs. Similar training speedup observations can be drawn from the example run in Figure~\ref{fig:baselines-comparison}. OpenAI baselines' PPO requires training of 100 minutes to solve Atari Pong, while rl\_games can tackle it within a fraction of time of the baseline, e.g., five minutes.

\section{Future Work}

\textbf{Completeness}: In this publication, we have only included RL environments with Atari~\cite{ALE}, MuJoCo~\cite{MuJoCo}, DeepMind Control Suite~\cite{dmc}, ViZDoom~\cite{VizDoom}, and classic ones like mountain car, cartpole, etc.
We intend to expand our pool of supported environments to cover more research use cases, e.g., grid worlds that are easily customized to research~\cite{minigrid}. 
On the multi-agent environments, we have implemented ViZDoom~\cite{VizDoom} and welcome the community to add even more environments including Google Research Football, MuJoCo Soccer Environment, etc.

\revision{\textbf{Cross-platform Support}: The \envpool~intends to support extra operating systems, such as MacOS and Windows.}

\revision{\textbf{User friendliness}: We intend to create a template repository to help customized environment integration into \envpool~easier, so that users can develop their own environment without having to work under \envpool's code base while still having access to register the self-written environment with \envpool~and use the \texttt{make} function to create it.}

\textbf{Distributed Use Case}: The \envpool~experiments in the paper have been performed on single machines. 
The same APIs can be extended to a distributed use case with remote execution of the environments using techniques like gRPC. The logic behind the environment execution is still hidden from the researchers but only the machines used to collect data will be at a much larger scale.

\textbf{Research Directions}: With such a high throughput of data generation, the research paradigm can be shifted to large-batch training to better leverage a large amount of generated data. 
There are no counterparts as successful in computer vision and natural language processing fields, where large-batch training leads to stable and faster convergence. 
An issue induced by faster environment execution would be severe off-policyness. Better off-policy RL algorithms are required to reveal the full power of the system.
Our proposed system also brings many new opportunities. For example, more accurate value estimations may be achieved by applying a large amount of parallel sampling, rollouts, and search.

\section{Conclusion}\label{section:conclusion}

In this work, we have introduced a highly parallel reinforcement learning environment execution engine \envpool, which significantly outperforms existing environment executors.
With a curated design dedicated to the RL use case, we leverage techniques of a general asynchronous execution model, implemented with a C++ thread pool on the environment execution. 
For data organization and outputting batch-wise observations, we designed BufferQueue tailored for the RL environments. 
We conduct an extensive study with various setups to demonstrate the scale-up ability of the proposed system and compare it with both the most popular implementation gym and highly optimized system Sample Factory. 
The conclusions hold for both Atari and MuJoCo, two of the most popular RL benchmark environments. 
In addition, we have demonstrated significant improvements in existing RL training libraries' speed when integrated with EnvPool in a wide variety of setups, including different machines, different RL environments, different RL algorithms, etc. On laptops with a GPU, we managed to train Atari Pong and MuJoCo Ant in five minutes, accelerating the development and iteration for researchers and practitioners. However, some limitations remain, for example, EnvPool cannot speed up RL environments originally written in Python and therefore developers have to translate each existing environment into C++. We hope that EnvPool will become a core component of modern RL infrastructures, providing easy speedup and high-throughput environment experiences for RL training systems.

\bibliographystyle{abbrv}
 
\bibliography{envpool}
\newpage
\section*{Checklist}

% %%% BEGIN INSTRUCTIONS %%%
% The checklist follows the references.  Please
% read the checklist guidelines carefully for information on how to answer these
% questions.  For each question, change the default \answerTODO{} to \answerYes{},
% \answerNo{}, or \answerNA{}.  You are strongly encouraged to include a {\bf
% justification to your answer}, either by referencing the appropriate section of
% your paper or providing a brief inline description.  For example:
% \begin{itemize}
%   \item Did you include the license to the code and datasets? \answerYes{See Section~.}
%   \item Did you include the license to the code and datasets? \answerNo{The code and the data are proprietary.}
%   \item Did you include the license to the code and datasets? \answerNA{}
% \end{itemize}
% Please do not modify the questions and only use the provided macros for your
% answers.  Note that the Checklist section does not count towards the page
% limit.  In your paper, please delete this instructions block and only keep the
% Checklist section heading above along with the questions/answers below.
% %%% END INSTRUCTIONS %%%

\begin{enumerate}

\item For all authors...
\begin{enumerate}
  \item Do the main claims made in the abstract and introduction accurately reflect the paper's contributions and scope?
    \answerYes{}
  \item Did you describe the limitations of your work?
    \answerYes{}
  \item Did you discuss any potential negative societal impacts of your work?
    \answerYes{}
  \item Have you read the ethics review guidelines and ensured that your paper conforms to them?
    \answerYes{}
\end{enumerate}

\item If you are including theoretical results...
\begin{enumerate}
  \item Did you state the full set of assumptions of all theoretical results?
    \answerNA{}
	\item Did you include complete proofs of all theoretical results?
    \answerYes{}
\end{enumerate}

\item If you ran experiments (e.g. for benchmarks)...
\begin{enumerate}
  \item Did you include the code, data, and instructions needed to reproduce the main experimental results (either in the supplemental material or as a URL)?
    \answerYes{}
  \item Did you specify all the training details (e.g., data splits, hyperparameters, how they were chosen)?
    \answerYes{}
	\item Did you report error bars (e.g., concerning the random seed after running experiments multiple times)?
    \answerYes{}
	\item Did you include the total amount of compute and the type of resources used (e.g., type of GPUs, internal cluster, or cloud provider)?
    \answerYes{}
\end{enumerate}

\item If you are using existing assets (e.g., code, data, models) or curating/releasing new assets...
\begin{enumerate}
  \item If your work uses existing assets, did you cite the creators?
    \answerYes{}
  \item Did you mention the license of the assets?
    \answerYes{}
  \item Did you include any new assets either in the supplemental material or as a URL?
    \answerYes{}
  \item Did you discuss whether and how consent was obtained from people whose data you're using/curating?
    \answerYes{}
  \item Did you discuss whether the data you are using/curating contains personally identifiable information or offensive content?
    \answerNA{}
\end{enumerate}

\item If you used crowdsourcing or conducted research with human subjects...
\begin{enumerate}
  \item Did you include the full text of instructions given to participants and screenshots, if applicable?
    \answerNA{}
  \item Did you describe any potential participant risks, with links to Institutional Review Board (IRB) approvals, if applicable?
    \answerNA{}
  \item Did you include the estimated hourly wage paid to participants and the total amount spent on participant compensation?
    \answerNA{}
\end{enumerate}

\end{enumerate}

%%%%%%%%%%%%%%%%%%%%%%%%%%%%%%%%%%%%%%%%%%%%%%%%%%%%%%%%%%%%

\appendix
%%%%%%%%%%%%%%%%%%%%%%%%%%%%%%%%%%%%%%%%%%%%%%%%%%%%%%%%%%%%

%%%%%%%%%%%%%%%%%%%%%%%%%%%%%%%%%%%%%%%%%%%%%%%%%%%%%%%%%%%%

\newpage
\appendix

\section{Usage of EnvPool}\label{appendix:python_apis}
In this section, we include comprehensive examples of the Python user APIs for EnvPool usage, including both synchronous and asynchronous execution modes, and for both OpenAI \texttt{gym} and \texttt{dm\_env} APIs.
\subsection{Synchronous  Execution, OpenAI \texttt{gym} APIs}
\begin{minted}[frame=single,framesep=10pt]{python}
import numpy as np
import envpool

# make gym env
env = envpool.make("Pong-v5", env_type="gym", num_envs=100)
obs = env.reset()  # with shape (100, 4, 84, 84)
act = np.zeros(100, dtype=int)
obs, rew, done, info = env.step(act, env_id=np.arange(100))
# env_id = info["env_id"]
\end{minted}

\subsection{Synchronous Execution, DeepMind \texttt{dm\_env} APIs}

\begin{minted}[frame=single,framesep=10pt]{python}
import numpy as np
import envpool

# make dm_env
env = envpool.make("Pong-v5", env_type="dm", num_envs=100)
obs = env.reset().observation.obs  # with shape (100, 4, 84, 84)
act = np.zeros(100, dtype=int)
timestep = env.step(act, env_id=np.arange(100))
# timestep.observation.obs, timestep.observation.env_id,
# timestep.reward, timestep.discount, timestep.step_type
\end{minted}

\subsection{Asynchronous Execution}
For maximizing the throughput of the environment execution, users may use the asynchronous execution mode. Both typical \texttt{step} API and more low-level APIs \texttt{recv}, \texttt{send} are provided. 
\begin{minted}[frame=single,framesep=10pt]{python}
import numpy as np
import envpool

# async by original API
env = envpool.make_dm("Pong-v5", num_envs=10, batch_size=9)
action_num = env.action_spec().num_values
timestep = env.reset()
env_id = timestep.observation.env_id
while True:
    action = np.random.randint(action_num, size=len(env_id))
    timestep = env.step(action, env_id)
    env_id = timestep.observation.env_id
\end{minted}
\newpage

\begin{minted}[frame=single,framesep=10pt]{python}
# or use low-level API, faster than previous version
env = envpool.make_dm("Pong-v5", num_envs=10, batch_size=9)
action_num = env.action_spec().num_values
env.async_reset()  # this can only call once at the beginning
while True:
    timestep = env.recv()
    env_id = timestep.observation.env_id
    action = np.random.randint(action_num, size=len(env_id))
    env.send(action, env_id)
\end{minted}

\begin{minted}[frame=single,framesep=10pt]{python}
import numpy as np
import envpool

# make asynchronous with gym API
num_envs = 10
batch_size = 9
env = envpool.make("Pong-v5", env_type="gym", num_envs=num_envs,
                   batch_size=batch_size)
env.async_reset()
while True:
    obs, rew, done, info = env.recv()
    env_id = info["env_id"]
    action = np.random.randint(batch_size, size=len(env_id))
    env.send(action, env_id)
\end{minted}

\section{CPU Specifications for Pure Environment Simulation}\label{appendix:cpu_type}
This section lists the detailed CPU specifications for the pure environment simulation experiments presented in the main paper.

The laptop has 12 Intel CPU cores, with Intel(R) Core(TM) i7-8750H CPU @ 2.20GHz. And the workstation has 32 AMD CPU cores, with AMD Ryzen 9 5950X 16-Core Processor. Evaluating \envpool~on these two configurations can demonstrate its effectiveness with small-scale experiments.

An NVIDIA DGX-A100 has 256 CPU cores with AMD EPYC 7742 64-Core Processor and 8 NUMA nodes. Note that running multi-processing on each NUMA node not only makes the memory closer to the processor but also reduces the thread contention on the \ABQ.

\section{Speed Improvements on Single Environment}\label{appendix:single_env}

We present experiments with a single environment (i.e., $N=1$) in Table~\ref{table:single_env}, where EnvPool manages to reduce overhead compared to the Python counterpart and achieves considerable speedup.

\begin{table}[tbh]
\centering
\caption{Single environment simulation speed on different hardware setups. The speed is in frames per second.} 
\label{table:single_env}

\begin{tabular}{c|c|c|c|c}
\toprule
System      & Method  & Atari Pong-v5 & MuJoCo Ant-v3 & dm\_control cheetah run \\ \midrule
Laptop      & Python  & 4,891      & 12,325      & 6,235                 \\
Laptop      & EnvPool & 7,887       & 15,641      & 11,636                \\
Laptop      & Speedup & 1.61$\times$         & 1.27$\times$        & 1.87$\times$                   \\ \midrule
Workstation & Python  & 7,739       & 19,472      & 9,042                 \\
Workstation & EnvPool & 12,623      & 25,725      & 16,691                \\
Workstation & Speedup & 1.63$\times$         & 1.32$\times$        & 1.85$\times$                   \\ \midrule
DGX-A100    & Python  & 4,449       & 11,018      & 5,024                 \\
DGX-A100    & EnvPool & 7,723       & 16,024      & 10,415                \\
DGX-A100    & Speedup & 1.74$\times$         & 1.45$\times$        & 2.07$\times$                   \\ \bottomrule 
\end{tabular}
\end{table}

\section{ActionBufferQueue and StateBufferQueue}\label{appendix:buffers}

\subsection{ActionBufferQueue}
\ABQ~is the queue that caches the actions from the \texttt{send} function, waiting to be consumed by the \TP. Many open-source general-purpose thread-safe event queues can be used for this purpose. In this work, we observe that in our case the total number of environments $N$, the \texttt{batch\_size} $M$, and the number of threads are all pre-determined at the construction of \envpool. The \ABQ~can thus be tailored for our specific case for optimal performance. 

We implemented \ABQ~with a lock-free circular buffer. A buffer with a size of $2N$ is allocated. We use two atomic counters to keep track of the head and tail of the queue. The counters modulo $2N$ is used as the indices to make the buffer circular. We use a \texttt{semaphore} to coordinate enqueue and dequeue operations and to make the threads wait when there is no action in the queue.

\subsection{StateBufferQueue}

\SBQ~is in charge of receiving data produced by each environment. Like the \ABQ, it is also tailored exactly for RL environments. \SBQ~is a lock-free circular buffer of memory blocks, each block contains a fixed number of slots equal to \texttt{batch\_size}, where each slot is for storing data generated by a single environment.

When one environment finishes its step inside \TP, the corresponding thread will acquire a slot in \SBQ~to write the data. When all slots are written, a block is marked as ready (see yellow slots in Figure~\ref{fig:sys}). By pre-allocating memory blocks, each block in \SBQ~can accommodate a batch of states. Environments will use slots of the pre-allocated space in a first come first serve manner. When a block is full, it can be directly taken as a batch of data, saving the overhead for batching. Both the allocation position and the write count of a block are tracked by atomic counters. When a block is ready, it is notified via a semaphore. Therefore the \SBQ~is also lock-free and highly performant.

\textbf{Data Movement} The popular Python vectorized environment executor performs memory copies at several places that are saved in \envpool. There is one inter-process copy for collecting the states from the worker processes, and one copy for batching the collected states. In \envpool, these copies are saved thanks to the \SBQ~ because:
\begin{itemize}
    \item We pre-allocate memory for a batch of states, the pointer to the target slot of memory is directly passed to the environment execution and written from the worker thread.
    \item The ownership of the block of memory is directly transferred to Python and converted into NumPy arrays via pybind11 when the block of memory is marked as ready.
\end{itemize}

\newpage

\section{Jitting for JAX}
Jitting the environment simulation code with the neural networks is supported via a set of jittable functions in EnvPool:

\begin{minted}[frame=single,framesep=10pt]{python}
import envpool
import jax.lax as lax

env = envpool.make(..., env_type="gym" | "dm")
handle, recv, send, step = env.xla()

def actor_step(iter, loop_var):
  handle0, states = loop_var
  action = policy(states)
  # for gym
  handle1, (new_states, rew, done, info) = step(handle0, action)
  # for dm
  # handle1, new_states = step(handle0, action)
  return (handle1, new_states)

@jit
def run_actor_loop(num_steps, init_var):
  return lax.fori_loop(0, num_steps, actor_step, init_var)

states = env.reset()
run_actor_loop(100, (handle, states))
\end{minted}

Currently, EnvPool supports jitting for CPU and GPU when used with JAX~\cite{jax2018github}. All jittable functions are implemented via XLA's custom call mechanism~\cite{customcall}. When the environment code is jitted, the control loop of the actor is lowered from Python code to XLA's runtime, allowing the entire control loop to run on a native thread and freeing the Python Global Interpreter Lock (GIL). Note that when the EnvPool function is jitted for GPU, the environment simulation is still executed on the CPU, as EnvPool only supports existing environments written in C/C++. We implemented GPU jitting by wrapping the CPU code in a GPU custom call, with memory transfers between two devices.

\section{Complete Results of End-to-end Agent Training} \label{appendix:end_to_end}

% \pagebreak
\subsection{CleanRL Training Results}
This section presents the complete training results using CleanRL's PPO and EnvPool. CleanRL's PPO closely matches the performance and implementation details of \texttt{openai/baselines}' PPO~\cite{shengyi2022the37implementation}. The source code is made available publicly\footnote{See \url{https://github.com/vwxyzjn/envpool-cleanrl}}. The hardware specifications for conducting the CleanRL's experiments are as follows:
\begin{itemize}
    \item OS: Pop!\_OS 21.10 x86\_64 
    \item Kernel: 5.17.5-76051705-generic 
    \item CPU: AMD Ryzen 9 3900X (24) @ 3.800GHz 
    \item GPU: NVIDIA GeForce RTX 2060 Rev. A 
    \item Memory: 64237MiB 
\end{itemize}

\begin{figure}[!htpb]
    \centering
    \includegraphics[width=1\linewidth]{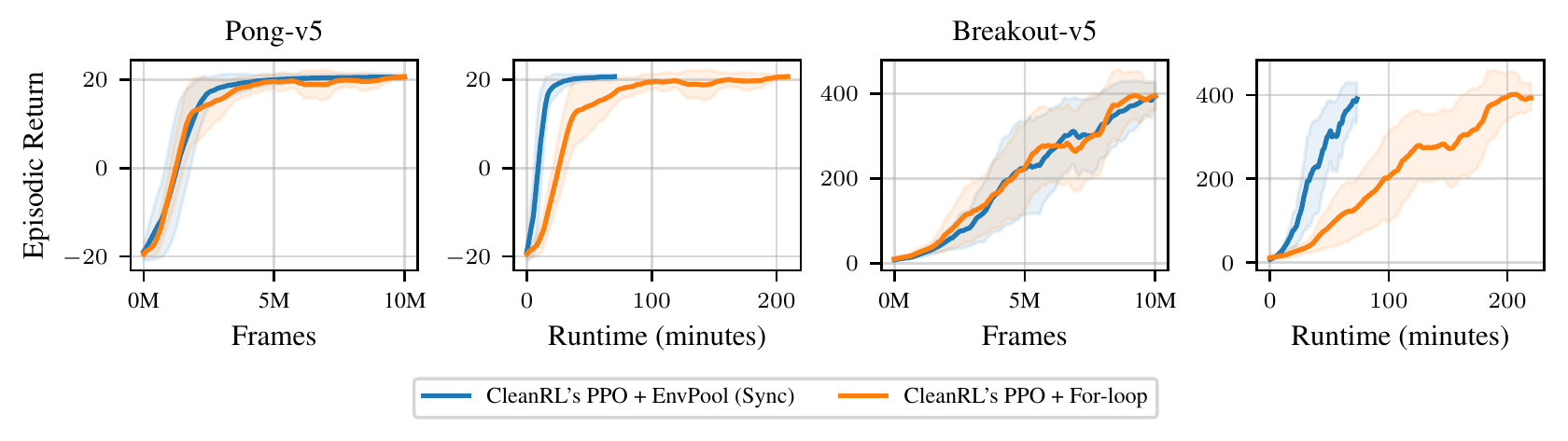}
    \caption{CleanRL example runs with Python vectorized environments and with EnvPool, using the same number of parallel environments $N=8$.}
    \label{fig:cleanrl-ppo-atari-learning-curves}
\end{figure}

\begin{figure}[!htpb]
    \centering
    \includegraphics[width=1\linewidth]{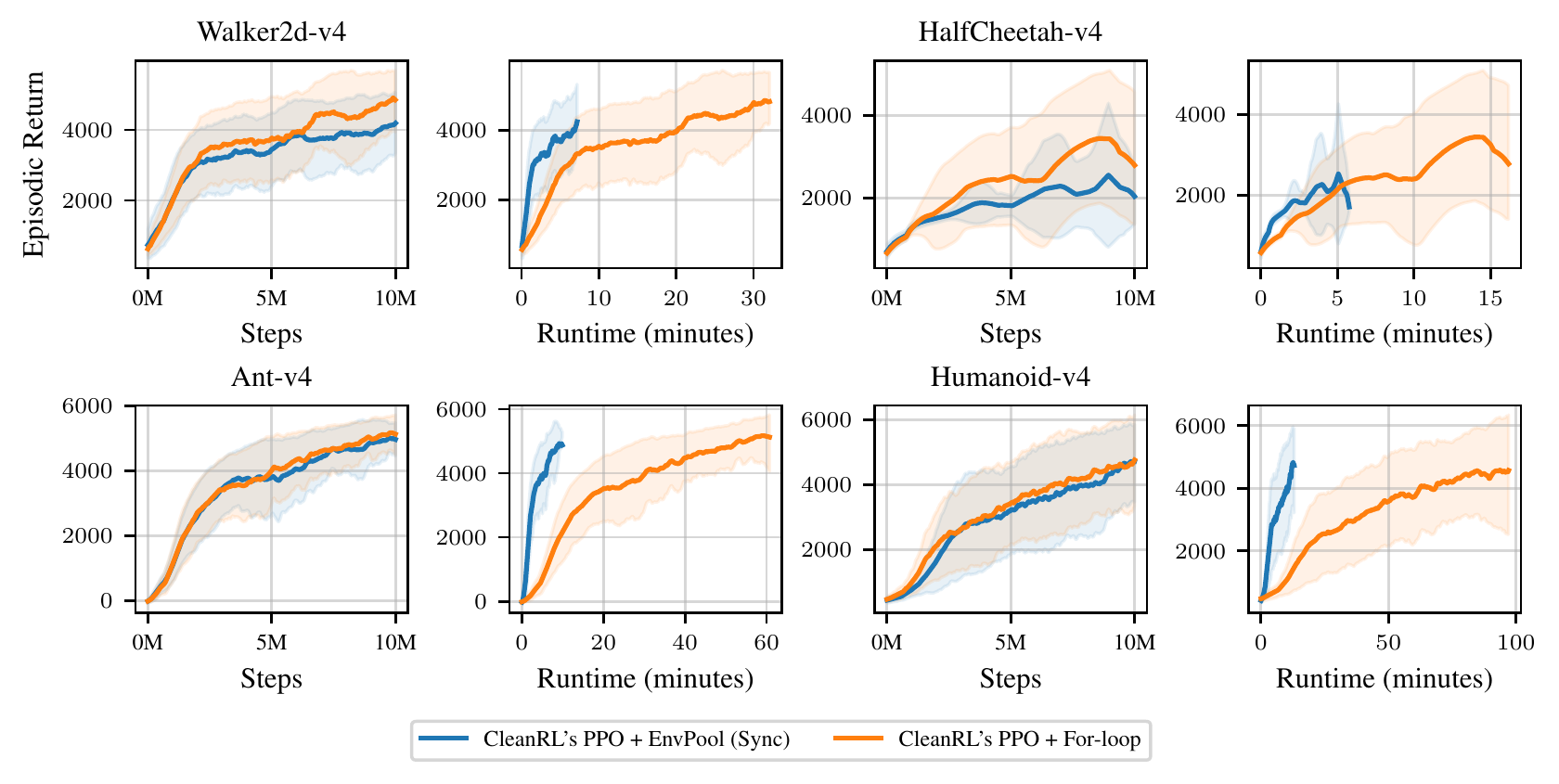}
        \caption{CleanRL example runs with Python vectorized environments and with EnvPool, using the same number of parallel environments $N=64$.}
        \label{fig:cleanrl-ppo-mujoco-learning-curves}
\end{figure}

CleanRL's Atari experiment's hyperparameters and learning curves can be found in  Table~\ref{tab:cleanrl-ppo-atari-params} and Figure~\ref{fig:cleanrl-ppo-atari-learning-curves}. CleanRL's MuJoCo experiment's hyperparameters and learning curves can be found in  Table~\ref{tab:cleanrl-ppo-mujoco-params} and Figure~\ref{fig:cleanrl-ppo-mujoco-learning-curves}. CleanRL's tuned Pong experiment's hyperparameters can be found in Table~\ref{tab:cleanrl-ppo-tuned-pong-params}.

Note the CleanRL's EnvPool experiments with MuJoCo use the \texttt{v4} environments and the gym's vecenv experiments use the \texttt{v2} environments. There are subtle differences between the \texttt{v2} and \texttt{v4} environments\footnote{See \url{https://github.com/openai/gym/pull/2762\#issuecomment-1135362092}}.

% Below are the hyperparameters
\begin{table}[ht]
\centering
\caption{PPO hyperparameters used for CleanRL's Atari experiments (i.e., \texttt{ppo\_atari.py} and \texttt{ppo\_atari\_envpool.py}). The hyperparameters used is aligned with \cite{PPO}. }
\begin{tabular}{ll} 
\toprule
Parameter Names  & Parameter Values\\
\midrule
$N_\text{total}$ Total Time Steps & 10,000,000  \\ 
$\alpha$ Learning Rate &  0.00025 Linearly Decreased to 0 \\
$N_\text{envs}$ Number of Environments & 8 \\
$N_\text{steps}$ Number of Steps per Environment & 128  \\
$\gamma$ (Discount Factor) & 0.99 \\ 
$\lambda$ (for GAE) & 0.95 \\ 
$N_\text{mb}$ Number of Mini-batches & 4 \\
$K$ (Number of PPO Update Iteration Per Epoch)& 4 \\
$\varepsilon$ (PPO's Clipping Coefficient) & 0.1 \\
$c_1$ (Value Function Coefficient)& 0.5\\
$c_2$ (Entropy Coefficient)& 0.01\\
$\omega$ (Gradient Norm Threshold)& 0.5 \\
Value Function Loss Clipping
\tablefootnote{See ``Value Function Loss Clipping'' in \cite{shengyi2022the37implementation}}
& True\\
\bottomrule
\end{tabular}
\label{tab:cleanrl-ppo-atari-params}
\end{table}
% Below are the hyperparameters

\begin{table}[ht]
\centering
\caption{PPO \emph{tuned} hyperparameters used for CleanRL's Pong experiments in Figure~\ref{fig:baselines-comparison}.}
\begin{tabular}{ll} 
\toprule
Parameter Names  & Parameter Values\\
\midrule
$N_\text{total}$ Total Time Steps & 7,000,000  \\ 
$\alpha$ Learning Rate &  0.002  \\
$N_\text{envs}$ Number of Environments & 64 \\
$N_\text{steps}$ Number of Steps per Environment & 128  \\
$\gamma$ (Discount Factor) & 0.99 \\ 
$\lambda$ (for GAE) & 0.95 \\ 
$N_\text{mb}$ Number of Mini-batches & 4 \\
$K$ (Number of PPO Update Iteration Per Epoch)& 4 \\
$\varepsilon$ (PPO's Clipping Coefficient) & 0.1 \\
$c_1$ (Value Function Coefficient)& 2.24 \\
$c_2$ (Entropy Coefficient)& 0.0 \\
$\omega$ (Gradient Norm Threshold)& 1.13 \\
Value Function Loss Clipping & False\\
\bottomrule
\end{tabular}
\label{tab:cleanrl-ppo-tuned-pong-params}
\end{table}

% The following learning curves come from the 

\begin{table}[ht]
\centering
\caption{PPO hyperparameters used for CleanRL's MuJoCo experiments (i.e., \texttt{ppo\_continuous\_action.py} and \texttt{ppo\_continuous\_action\_envpool.py}). Note that \cite{PPO} uses $N_\text{envs} = 1$ so we needed to find an alternative set of hyperparameters.}
\begin{tabular}{ll} 
\toprule
Parameter Names  & Parameter Values\\
\midrule
$N_\text{total}$ Total Time Steps & 10,000,000  \\ 
$\alpha$ Learning Rate &  0.00295 Linearly Decreased to 0 \\
$N_\text{envs}$ Number of Environments & 64 \\
$N_\text{steps}$ Number of Steps per Environment & 64  \\
$\gamma$ (Discount Factor) & 0.99 \\ 
$\lambda$ (for GAE) & 0.95 \\ 
$N_\text{mb}$ Number of Mini-batches & 4 \\
$K$ (Number of PPO Update Iteration Per Epoch)& 2 \\
$\varepsilon$ (PPO's Clipping Coefficient) & 0.2 \\
$c_1$ (Value Function Coefficient)& 1.3 \\
$c_2$ (Entropy Coefficient)& 0.0\\
$\omega$ (Gradient Norm Threshold)& 3.5 \\
Value Function Loss Clipping & False \\
\bottomrule
\end{tabular}
\label{tab:cleanrl-ppo-mujoco-params}
\end{table}

\clearpage

\subsection{rl\_games Training Results}

This section presents the complete training results using rl\_games' PPO and EnvPool. The hyperparameters configuration can be found in rl\_games' repository.\footnote{See the files postfixed with \texttt{envpool} in \url{https://github.com/Denys88/rl_games/tree/master/rl_games/configs/atari}.} The hardware specifications for conducting the rl\_games' experiments are as follows:
\begin{itemize}
    \item OS: Ubuntu 21.10 x86\_64
    \item Kernel: 5.13.0-48-generic
    \item CPU: 11th Gen Intel i9-11980HK (16) @ 4.900GHz
    \item GPU: NVIDIA GeForce RTX 3080 Mobile / Max-Q 8GB/16GB
    \item Memory: 64,012MiB
\end{itemize}

The commit used to run the experiments is \href{https://github.com/Denys88/rl_games/tree/7f259a6436f396274c9931d0bd7004cee2ecabfa}{\texttt{7f259a6436f396274c9931d0bd7004cee2ecabfa}}, and the hyperparameters are tuned per environment and are available at
\begin{itemize}
    \item Ant: \href{https://github.com/Denys88/rl_games/blob/7f259a6436f396274c9931d0bd7004cee2ecabfa/rl_games/configs/MuJoCo/ant_envpool.yaml}{rl\_games/configs/MuJoCo/ant\_envpool.yaml}
    \item Walker2D: \href{https://github.com/Denys88/rl_games/blob/7f259a6436f396274c9931d0bd7004cee2ecabfa/rl_games/configs/MuJoCo/walker2d_envpool.yaml}{rl\_games/configs/MuJoCo/walker2d\_envpool.yaml}
    \item HalfCheetah: \href{https://github.com/Denys88/rl_games/blob/7f259a6436f396274c9931d0bd7004cee2ecabfa/rl_games/configs/MuJoCo/halfcheetah_envpool.yaml}{rl\_games/configs/MuJoCo/halfcheetah\_envpool.yaml}
    \item Humanoid \href{https://github.com/Denys88/rl_games/blob/7f259a6436f396274c9931d0bd7004cee2ecabfa/rl_games/configs/MuJoCo/humanoid_envpool.yaml}{rl\_games/configs/MuJoCo/humanoid\_envpool.yaml}
    \item Breakout \href{https://github.com/Denys88/rl_games/blob/7f259a6436f396274c9931d0bd7004cee2ecabfa/rl_games/configs/atari/ppo_breakout_envpool.yaml}{rl\_games/configs/atari/ppo\_breakout\_envpool.yaml}
    \item Pong \href{https://github.com/Denys88/rl_games/blob/7f259a6436f396274c9931d0bd7004cee2ecabfa/rl_games/configs/atari/ppo_pong_envpool.yaml}{rl\_games/configs/atari/ppo\_pong\_envpool.yaml}
\end{itemize}

\begin{table}[ht]
\centering
\caption{PPO baseline hyperparameters used for rl\_games's MuJoCo experiments. Some environments use different neural network architectures.}
\begin{tabular}{ll} 
\toprule
Parameter Names  & Parameter Values\\
\midrule
$N_\text{envs}$ Number of Environments & 64 \\
$N_\text{steps}$ Number of Steps per Environment & 256 for HalfCheetah \\
 & 64 for Ant \\
 & 128 for Humanoid and Walker2D \\
$\gamma$ (Discount Factor) & 0.99 \\ 
$\lambda$ (for GAE) & 0.95 \\ 
$N_\text{mb}$ Number of Mini-batches & 2 \\
$K$ (Number of PPO Update Iteration Per Epoch)& 4 for Ant \\
 & 5 for HalfCheetah, Walker2D and Humanoid  \\
$\varepsilon$ (PPO's Clipping Coefficient) & 0.2 \\
$c_1$ (Value Function Coefficient)& 2.0 \\
$c_2$ (Entropy Coefficient)& 0.0 \\
$\omega$ (Gradient Norm Threshold)& 1.0 \\
$\alpha$ Learning Rate &  0.0003 dynamically adapted based on $\nu$  \\
$\nu$ KL Divergence threshold (for $\alpha$) &  0.008  \\
Value Function Loss Clipping & True \\
Value Bootstrap on Terminal States & True  \\
Reward Scale & 0.1  \\
Smooth Ratio Clamp & True  \\
Observation Normalization & True \\
Value Normalization & True  \\
MLP Sizes & $[128, 64, 32]$ for HalfCheetah \\
 & $[256, 128, 64]$ for Ant and Walker2D \\
 & $[512, 256, 128]$ for Humanoid \\
MLP Activation & Elu \\
Shared actor critic network & True \\
MLP Layer Initializer & Xavier \\
\bottomrule
\end{tabular}
\label{tab:rl-games-ppo-ant-params}
\end{table}

We compare the training performance of rl\_games using EnvPool against using Ray~\cite{Ray}'s parallel environments. In Figure~\ref{fig:rl-games-ppo-MuJoCo-learning-curves}, it's observed that EnvPool can boost the training system with multiple times training speed compared to Ray's integration. 

For example using well-tuned hyperparameters we can train Atari Pong game in under 2 min to 18+ training and 20+ evaluation score on a laptop.

\begin{figure}[!htpb]
    \centering
    \includegraphics[width=1\linewidth]{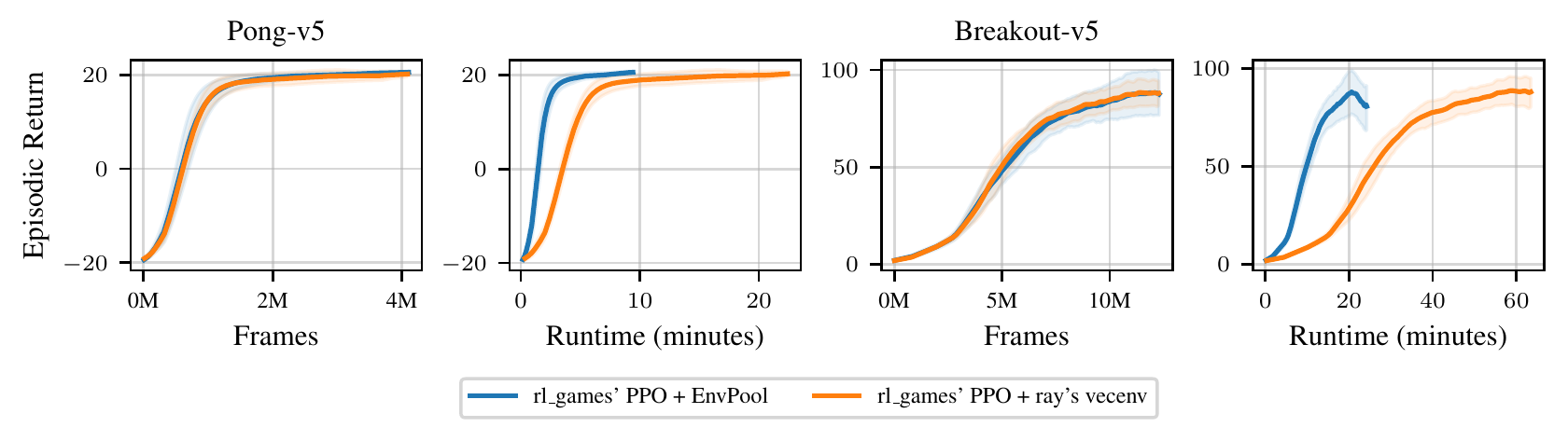}
    \caption{rl\_games example runs with Ray environments and with EnvPool, using the same number of parallel environments $N=64$.}
    \label{fig:rl-games-ppo-atari-learning-curves}
\end{figure}

\begin{figure}[tbh]
    \centering
    \includegraphics[width=1\linewidth]{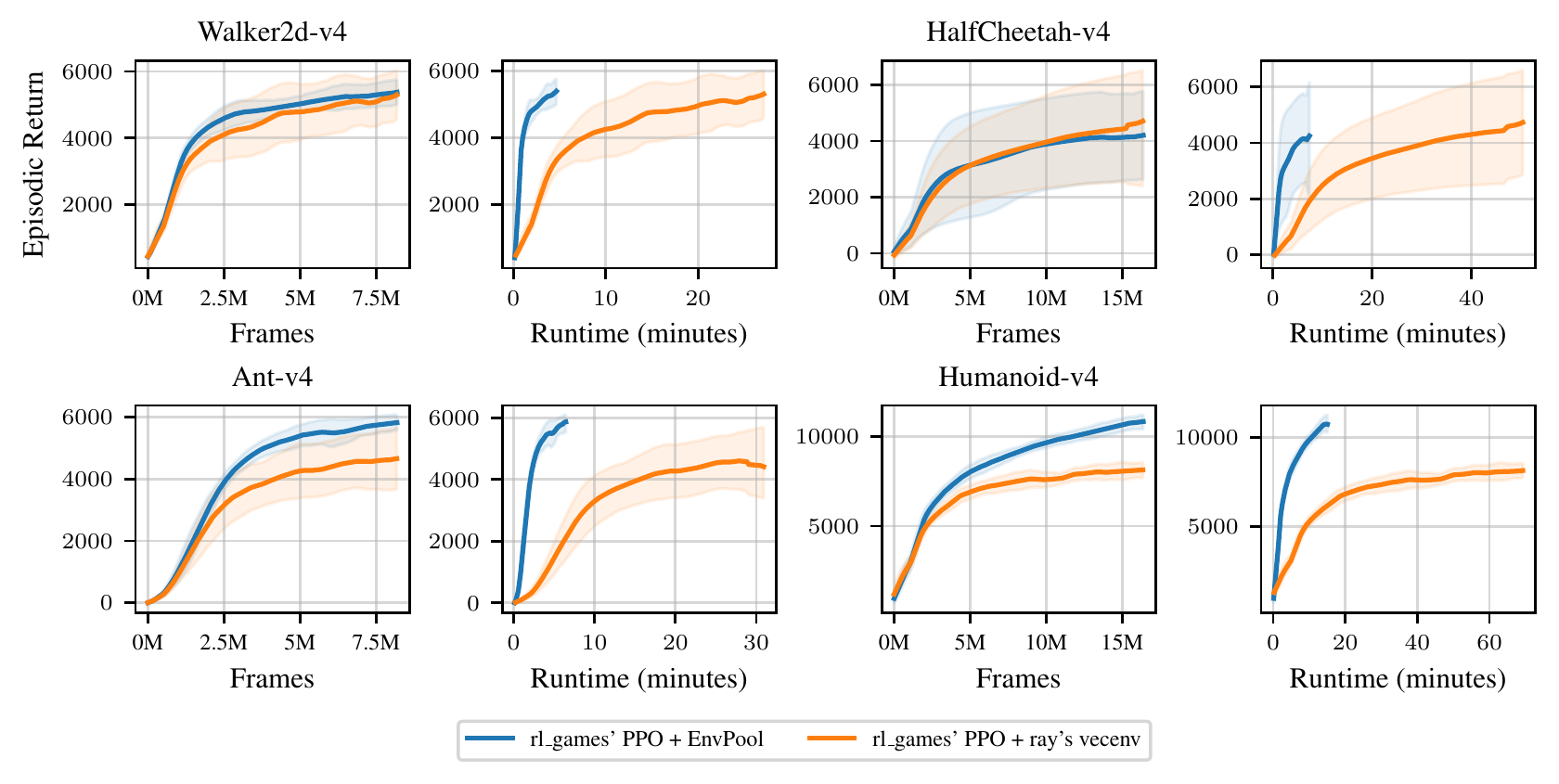}
        \caption{rl\_games example runs with Ray environments and with EnvPool, using the same number of parallel environments $N=64$.}
        \label{fig:rl-games-ppo-MuJoCo-learning-curves}
\end{figure}

Well-established implementations of SAC~\cite{huang2021cleanrl,tianshou} can only train Humanoid to a score of 5,000 in three to four hours. It's worth highlighting that we can now train Humanoid to a score over 10,000 just in 15 minutes with a laptop.

rl\_games's Atari experiment's learning curves can be found in  and Figure~\ref{fig:rl-games-ppo-atari-learning-curves}. rl\_games's MuJoCo experiment's learning curves can be found in Figure~\ref{fig:rl-games-ppo-MuJoCo-learning-curves}. Table~\ref{tab:rl-games-ppo-ant-params}, Table~\ref{tab:rl-games-ppo-atari-params} and Table~\ref{tab:rl-games-ppo-atari-pong-params} are the hyperparameters.

\begin{table}[ht]
\centering
\caption{PPO hyperparameters used for rl\_games's Atari Breakout experiments. }
\begin{tabular}{ll} 
\toprule
Parameter Names  & Parameter Values\\
\midrule
$N_\text{envs}$ Number of Environments & 64 \\
$N_\text{steps}$ Number of Steps per Environment & 128  \\
$\gamma$ (Discount Factor) & 0.999 \\ 
$\lambda$ (for GAE) & 0.95 \\ 
$N_\text{mb}$ Number of Mini-batches & 4 \\
$K$ (Number of PPO Update Iteration Per Epoch)& 2 \\
$\varepsilon$ (PPO's Clipping Coefficient) & 0.2 \\
$c_1$ (Value Function Coefficient)& 2.0 \\
$c_2$ (Entropy Coefficient)& 0.01 \\
$\omega$ (Gradient Norm Threshold)& 1.0 \\
$\alpha$ Learning Rate &  0.0008 \\
Value Function Loss Clipping & False \\
Observation Normalization & False \\
Value Normalization & True  \\
Neural network & Nature CNN \\
Activation & ReLU \\
Shared actor critic network & True \\
Layer Initializer & Orthogonal \\
\bottomrule
\end{tabular}
\label{tab:rl-games-ppo-atari-params}
\end{table}

\begin{table}[ht]
\centering
\caption{PPO hyperparameters used for rl\_games's Atari Pong experiments. }
\begin{tabular}{ll} 
\toprule
Parameter Names  & Parameter Values\\
\midrule
$N_\text{total}$ Total Time Steps & 8,000,000  \\ 
$N_\text{envs}$ Number of Environments & 64 \\
$N_\text{steps}$ Number of Steps per Environment & 128  \\
$\gamma$ (Discount Factor) & 0.999 \\ 
$\lambda$ (for GAE) & 0.95 \\ 
$N_\text{mb}$ Number of Mini-batches & 8 \\
$K$ (Number of PPO Update Iteration Per Epoch)& 2 \\
$\varepsilon$ (PPO's Clipping Coefficient) & 0.2 \\
$c_1$ (Value Function Coefficient)& 2.0 \\
$c_2$ (Entropy Coefficient)& 0.01 \\
$\omega$ (Gradient Norm Threshold)& 1.0 \\
$\alpha$ Learning Rate &  0.0003 dynamically adapted based on $\nu$  \\
$\nu$ KL Divergence threshold (for $\alpha$) &  0.01  \\
Value Function Loss Clipping & True \\
Observation Normalization & True \\
Value Normalization & True  \\
Neural network & Nature CNN \\
Activation & Elu \\
Shared actor critic network & True \\
Layer Initializer & Orthogonal \\
\bottomrule
\end{tabular}
\label{tab:rl-games-ppo-atari-pong-params}
\end{table}

\clearpage
\subsection{Acme-based Training Results}
We integrate EnvPool with Acme~\citep{acme} for experiments of PPO~\citep{PPO} in MuJoCo tasks, to show EnvPool's efficiency with different \texttt{num\_envs} and its advantage over other vectorized environments such as Stable Baseline's \textit{DummyVecEnv}~\citep{stable-baselines3}. The codes and hyperparameters can be found in our open-sourced codebase.

All the experiments were performed on a standard TPUv3-8 machine on Google Cloud with the following hardware specifications:

\begin{itemize}
    \item OS: Ubuntu 20.04 x86\_64
    \item Kernel: 5.4.0-1043-gcp
    \item CPU: Intel(R) Xeon(R) CPU @ 2.00GHz
    \item TPU: v3-8 with v2-alpha software
    \item Memory: 342,605MiB
\end{itemize}

\begin{figure}[!htpb]
    \centering
    \includegraphics[width=0.5\linewidth]{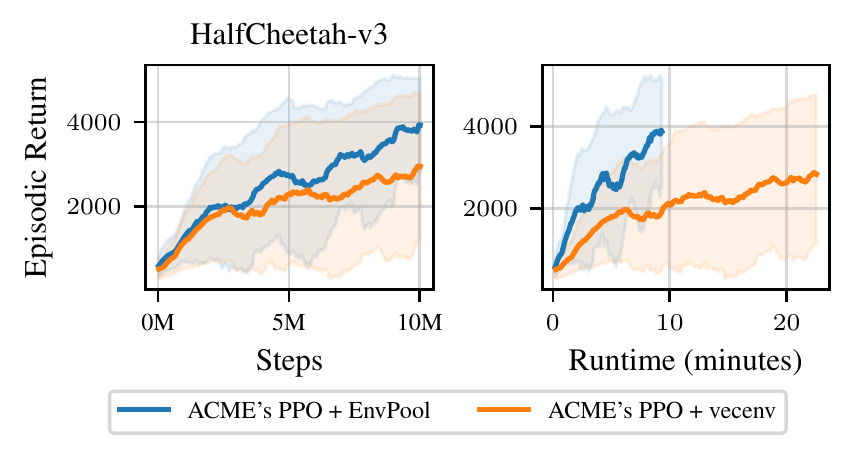}
    \caption{Comparison of EnvPool and DummyVecEnv using Acme's PPO implementation on MuJoCo HalfCheetah-v3 environment. Both settings use \texttt{num\_envs} of 32.}
    \label{fig:acme-vec-env-comparison}
\end{figure}

In Figure~\ref{fig:acme-vec-env-comparison}, we compare EnvPool with another popular batch environment \textit{DummyVecEnv}, which is argued to be more efficient than its alternative \textit{SubprocVecEnv} for light environments, both provided by Stable Baseline 3 \citep{stable-baselines3}. Using the same number of environments, EnvPool spent less than half of the wall time of \textit{DummyVecEnv} to achieve a similar final episode return, proving our efficiency.

\begin{figure}[!htpb]
    \centering
    \includegraphics[width=0.5\linewidth]{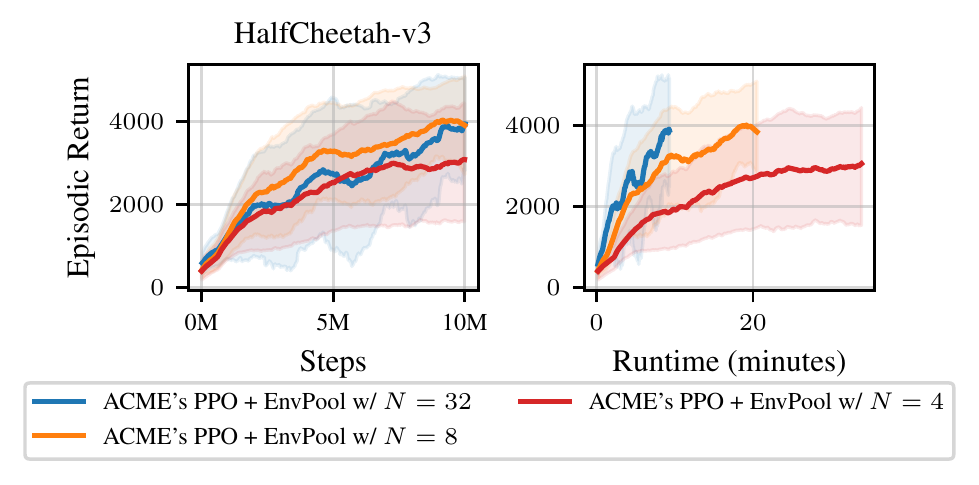}
    \caption{Training curves of Acme's PPO implementation on MuJoCo HalfCheetah-v3 environment using EnvPool of a different number of parallel environments.}
    \label{fig:acme-diff-num-envs}
\end{figure}

Figure~\ref{fig:acme-diff-num-envs} shows that under the same environment interaction budget, tuning the \texttt{num\_envs} can greatly reduce the training time while maintaining similar sample efficiency. We note that the key to maintaining the sample efficiency is to keep the same amount of environment steps under the same set of policy parameters. In our case, we simply maintain $\texttt{num\_envs} \times \texttt{batch\_size}$ a constant.

\pagebreak

\newpage

\section{Hyper-parameters Tuning} \label{appendix:hyperparams-tuning}
The motivation to find a set of hyperparameters is well-explained in Section~\ref{sec:end-to-end-training} — the idea is to use a large $N$ such as 32 or 64 to better leverage EnvPool's throughput and use less stale data.  This section explains the process of tuning hyperparameters in this paper.

Specifically, regarding the CleanRL + Atari experiments in Figure~\ref{fig:case-study}, we have used the same hyperparameters as in the original PPO paper~\citep[Table 5]{PPO}. CleanRL’s hyperparameters in Figure~\ref{fig:cleanrl-ppo-mujoco-learning-curves} were tuned via Weights and Biases’ automated hyperparameters search that optimizes average normalized scores in HalfCheetah-v4, Walker2d-v4, and Ant-v4 for 3 random seeds using Bayesian optimization. CleanRL’s hyperparameters in Table~\ref{tab:cleanrl-ppo-tuned-pong-params} were tuned via a similar procedure to optimize just for Pong-v5.

All the hyperparameters used rl\_games were tuned through trial and error, following the same practice in IsaacGym~\citep{isaac-gym}.

\section{License of EnvPool and RL Environments}

EnvPool is under Apache2 license. Other third-party source codes and data are under their corresponding licenses.

\section{Data Collection Process and Broader Social Impact}
All the data outputted by EnvPool is generated by the underlying simulators and game engines. Wrappers (e.g., transformation of the inputs) implemented in EnvPool conduct data pre-processing for the learning agents. EnvPool provides an effective way to parallel execution of the RL environments and does not have the typical supervised learning data labelling or collection process.

EnvPool is an infrastructure component to improve the throughput of generating RL experiences and the training system. EnvPool does not change the nature of the underlying RL environments or the training systems. Thus, to the best of the authors' knowledge, EnvPool does not introduce extra social impact to the field of RL and AI apart from our technical contributions.
\section{Author Contributions}
\label{appendix:author}

% The following individuals contributed to the development and implementation of EnvPool:

\textbf{Jiayi Weng and Min Lin} designed and implemented the core infrastructure of EnvPool.

\textbf{Shengyi Huang} originally demonstrated the effectiveness of end-to-end agent training with Atari Pong and Breakout using EnvPool.

\textbf{Jiayi Weng} conducted pure environment simulation experiments.

\textbf{Bo Liu} conducted environment alignment test and contributed to EnvPool bug reports and debugging.

\textbf{Shengyi Huang, Denys Makoviichuk, Viktor Makoviychuk, and Zichen Liu} conducted end-to-end agent training experiments with CleanRL, rl\_games, Ray, and Acme with Atari and MuJoCo environments.

\textbf{Jiayi Weng} developed Atari, ViZDoom, and OpenAI Gym Classic Control and Toy Text environments in EnvPool.

\textbf{Jiayi Weng, Bo Liu, and Yufan Song} developed OpenAI Gym MuJoCo and DeepMind Control Suite benchmark environments in EnvPool.

% \textbf{Bo Liu} developed Google Research Football environments in EnvPool.

\textbf{Jiayi Weng and Ting Luo} developed OpenAI Gym Box2D environments in EnvPool.

\textbf{Yukun Jiang} developed OpenAI Procgen environments in EnvPool.

\textbf{Shengyi Huang} implemented CleanRL's PPO integration with EnvPool.

\textbf{Denys Makoviichuk and Viktor Makoviychuk} implemented rl\_games integration with EnvPool.

\textbf{Zichen Liu} contributed to EnvPool Acme integration.

\textbf{Min Lin and Zhongwen Xu} led the project from its inception.

\textbf{Shuicheng Yan} advised and supported the project.

\textbf{Jiayi Weng, Zhongwen Xu, Min Lin, Shengyi Huang, Bo Liu, Denys Makoviichuk, and Viktor Makoviychuk} wrote the paper.

\end{document}